\documentclass{cta-author}

{}
{}
{}
\usepackage{float}
\usepackage{color}

\begin{document}

\supertitle{Submission Template for IET Research Journal Papers}

\title{Crowd counting with segmentation attention convolutional neural network}

\author{\au{Jiwei Chen$^{1,2}$}, \au{Zengfu Wang $^{1,2\corr}$}}

\address{\add{1}{Institute of Intelligent Machines, Chinese Academy of Sciences, Hefei, Anhui, People's Republic of China}
\add{2}{Department of Automation, University of Science and Technology of China, Hefei, Anhui, People's Republic of China}
\email{zfwang@ustc.edu.cn}}


\begin{abstract}
Deep learning occupies an undisputed dominance in crowd counting. In this paper, we propose a novel convolutional neural network (CNN) architecture called SegCrowdNet. Despite the complex background in crowd scenes, the proposed SegCrowdNet still adaptively highlights the human head region and suppresses the non-head region by segmentation. With the guidance of an attention mechanism, the proposed SegCrowdNet pays more attention to the human head region and automatically encodes the highly refined density map. The crowd count can be obtained by integrating the density map. To adapt the variation of crowd counts, SegCrowdNet intelligently classifies the crowd count of each image into several groups. In addition, the multi-scale features are learned and extracted in the proposed SegCrowdNet to overcome the scale variations of the crowd. To verify the effectiveness of our proposed method, extensive experiments are conducted on four challenging datasets. The results demonstrate that our proposed SegCrowdNet achieves excellent performance compared with the state-of-the-art methods.
\end{abstract}

\maketitle

\section{Introduction}\label{sec1}


Recently, crowd counting has become a research hotspot owing to its wide applications including public safety management \cite{gao2017crowd}, crowd analysis \cite{ma2018scene}, and urban planning \cite{zhan2008crowd}. Besides, the methods of crowd counting have important references in vehicle counting, cell counting, and other objects counting \cite{marsde2018people}. However, due to the crowd's complexity such as severe occlusion, high diversity, scale variation, view-point variation, and non-uniform distribution, the accuracy of crowd counting still has significant room for improvement. 

  Lots of efforts have been done in crowd counting. We generally classify these methods into detection-based methods and regression-based methods. The detection-based methods \cite{dalal2005histograms,pham2015count,viola2004robust,benenson2014ten} obtain the crowd count by counting the number of positive detections. The regression-based methods \cite{chan2009bayesian,lempitsky2010learning,idrees2013multi,pham2015count} utilize the mapping between extracted features and cropped patches to regress the crowd count. However, the aforementioned methods are complex, because their features need to be extracted by extremely numerous and cumbersome designs.
  In recent years, deep learning has obtained breakthroughs in many fields \cite{walach2016learning,long2015fully,ren2015faster}. Many CNN-based methods \cite{zhang2016single,sam2017switching,sindagi2017generating} have been proposed in crowd counting. They learn the non-linear mapping between training images and crowd counts. Researchers employ these CNN models to generate a density map that records the count and spatial information of the crowd at each pixel location, then the density map is integrated to get the crowd count.

\begin{figure}[t]
\setlength{\abovecaptionskip}{0.001cm}
\begin{center}
   \includegraphics[width=1.0\linewidth]{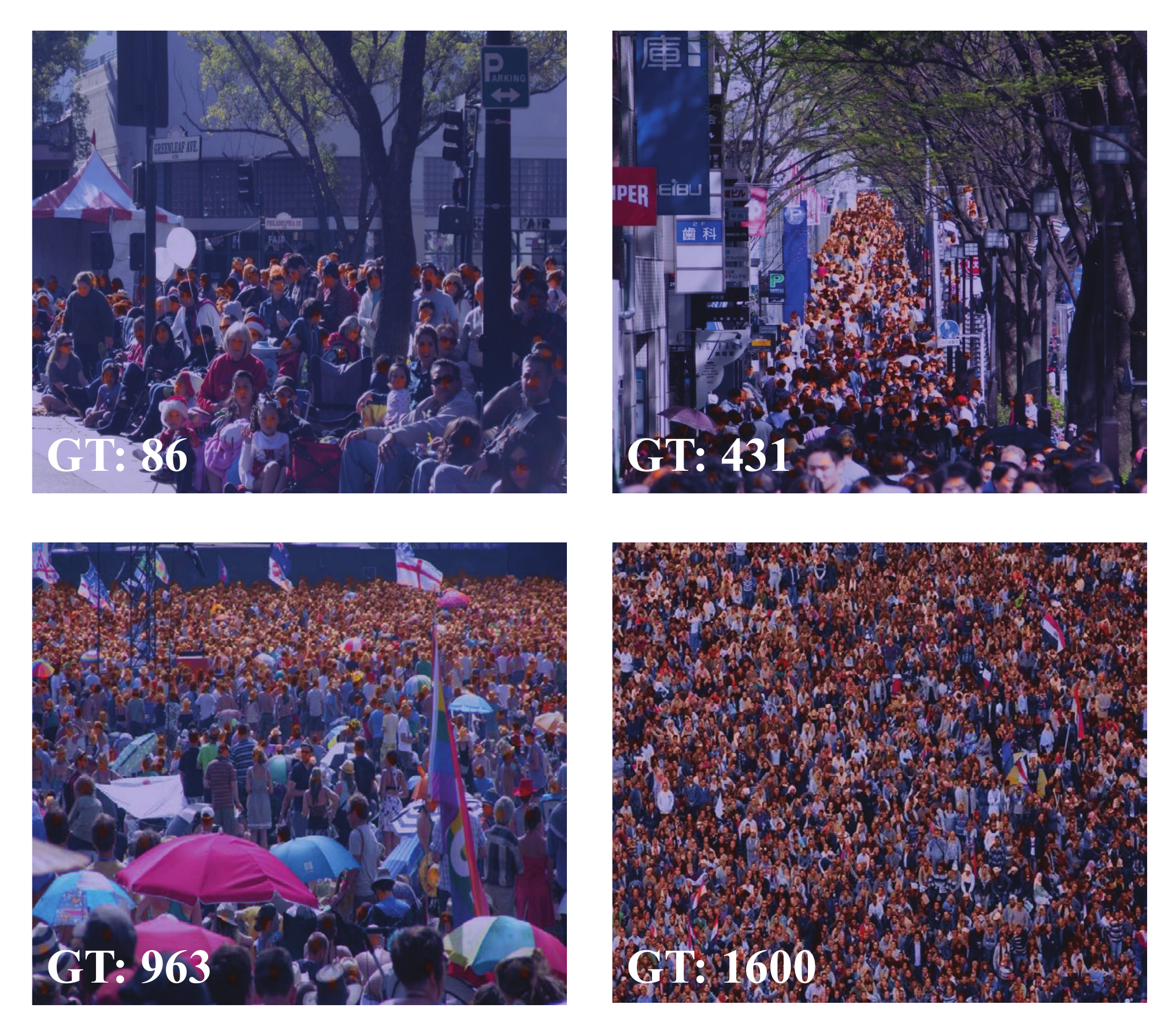}
\end{center}
   \caption{Two challenges for crowd counting. (1) Only the human head (highlight area) belongs to the foreground. The background is extremely complex. (2) The crowd count in each image varies greatly. GT represents the crowd count of the image.}
\label{fig:long}
\label{fig:onecol}
\end{figure}

Although deep learning has achieved great improvement in crowd counting, we find that the background is extremely complex in the crowd scene, and we believe that the accuracy of crowd counting can be further improved with the segmentation attention mechanism which highlights the foreground, suppresses the background and makes the proposed SegCrowdNet pay more attention to the foreground. Some example images are shown in Fig. 1. Let's carefully distinguish the foreground and background in crowd counting. We obtain the crowd counting result by counting the number of the human head. The human head belongs to the foreground. Everything apart from the human head belongs to the background. Therefore, the background is extremely complicated. In Fig. 1, we can also observe that the crowd count changes dramatically across images, which is the second challenging problem. To address these two challenging problems, firstly, we predict a segmentation map to adaptively highlight the human head region and suppress the non-head region. With the guidance of the segmentation results, the adaptive attention weights are used in the estimation of the density map to guide our network called SegCrowdNet to pay more attention to the human head region and generate a highly refined density map. Integrating the refined density map can get the crowd counting result. Unfortunately, the present datasets don't provide the ground truth of segmentation. We propose a simple but effective method in which the ones template is pasted on a binary map to encode the ground truth of segmentation. Secondly, to adapt the large variation of crowd counts, we utilized a classification task where the crowd count of each image is classified automatically. Meanwhile, to extract the multi-scale features to adapt the multi-scale crowd, our proposed SegCrowdNet not only utilizes different convolution kernels to encode the image but also fuses rich hierarchies from different depths of convolutional layers where the lower layer can extract discriminative features of the pedestrian and the higher layer can learn semantic concepts of the same pedestrian.

In this paper, we propose a novel end-to-end framework called SegCrowdNet. To the best of our knowledge, the proposed SegCrowdNet is the first network to utilize the segmentation attention mechanism in crowd counting. It adaptively highlights the human head region and suppresses the non-head region by way of optimizing a novel loss. With the guidance of the segmentation attention mechanism, the proposed SegCrowdNet pays more attention to the human head region and automatically encodes the highly refined density map. Our proposed SegCrowdNet can also automatically adapt the variation of crowd counts by learning a classification function. Extensive experiments are conducted on ShanghaiTech Part\_A dataset \cite{zhang2016single}, ShanghaiTech Part\_B dataset \cite{zhang2016single}, UCF\_CC\_50 dataset \cite{idrees2013multi} and  WorldExpo’10 dataset \cite{zhang2015cross}. The results demonstrate that our proposed method outperforms many state-of-the-art methods.

\begin{figure*}[t]
\begin{center}
   \includegraphics[width=1.0\linewidth]{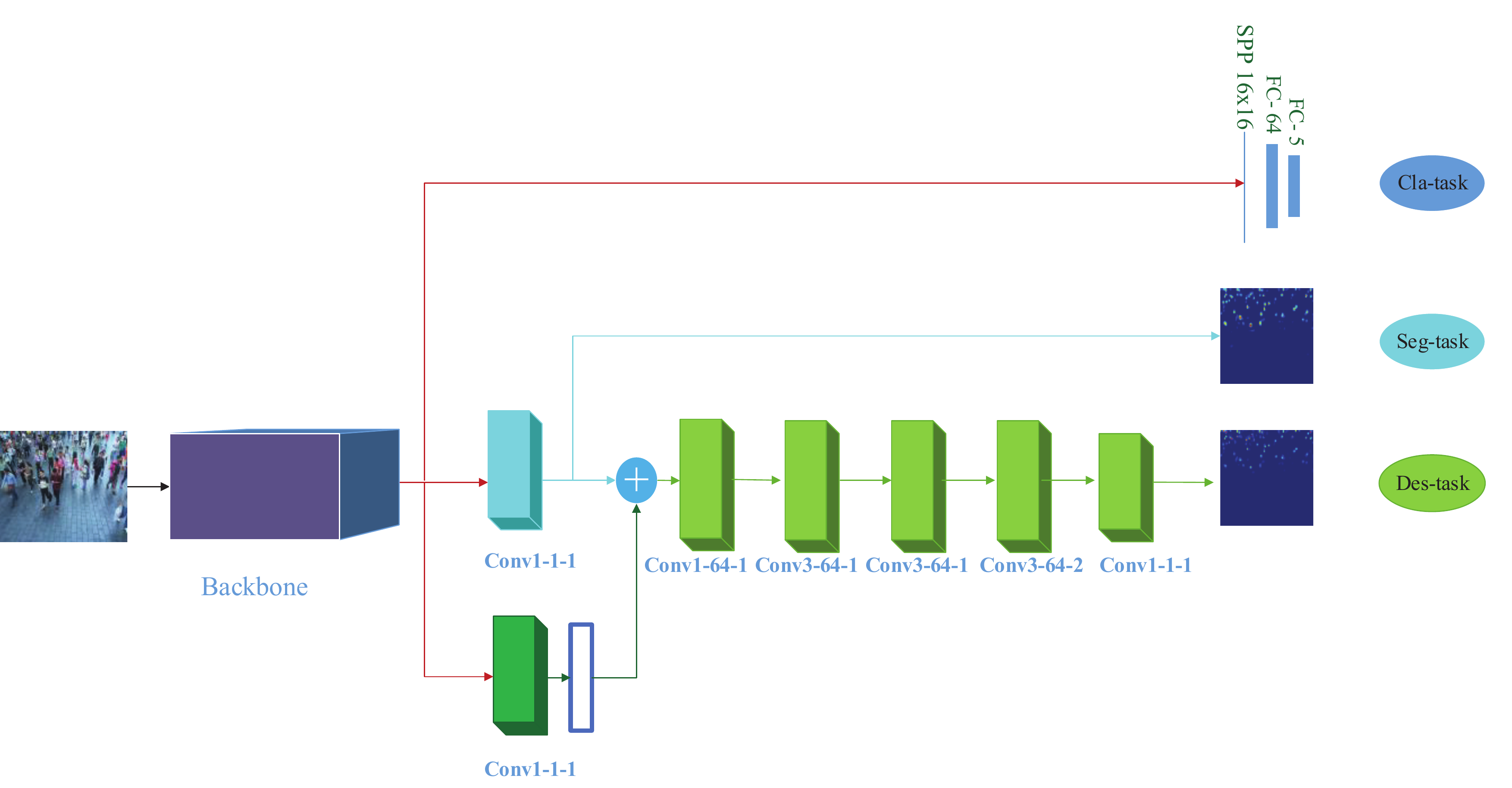}
\end{center}
   \caption{The proposed architecture of our SegCrowdNet. It is a multi-task model including the classification task (Cla-task), segmentation task (Seg-task), and density estimation task (Des-task). The convolutional layers’
parameters are represented as “Con(kernel size)-(number of filters)-
(dilation rate)”. $\bigoplus$ represents the element-wise add. The estimated density map (outlined in blue) is encoded in the intermediate supervision process.}
\label{fig:long}
\label{fig:onecol}
\end{figure*}

\begin{figure*}[t]
\begin{center}
   \includegraphics[width=1.0\linewidth]{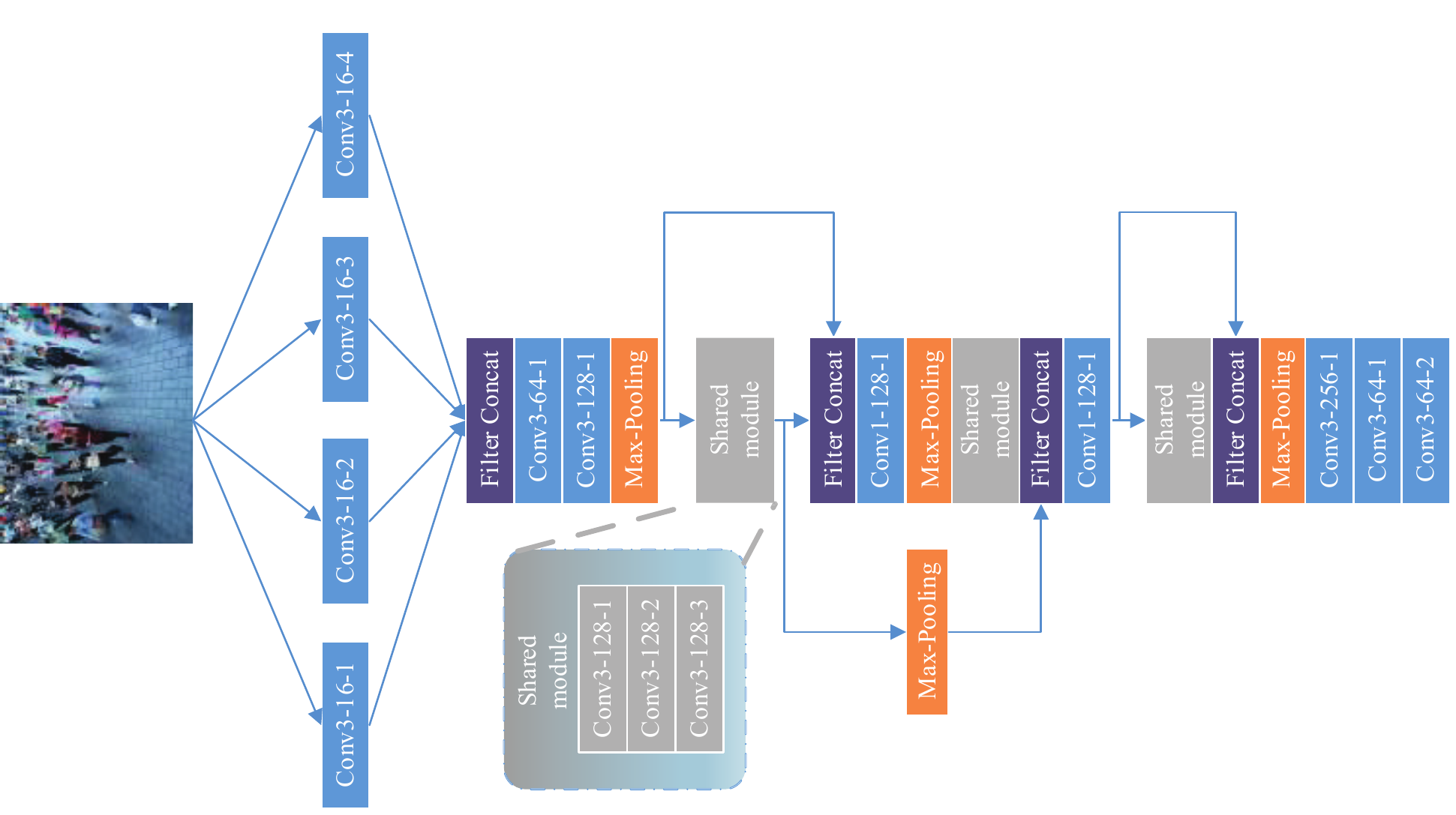}
\end{center}
   \caption{The proposed architecture of the backbone. The convolutional layers’
parameters are represented as “Conv(kernel size)-(number of filters)-
(dilation rate)”. The kernel size of the max pool is $2\times 2$ with stride 2.}
\label{fig:long}
\label{fig:onecol}
\end{figure*}

\section{Related work}
The researches \cite{sindagi2018survey,li2015crowded,Xu2019,Sajid2017Crowd,Mahmood2017,SivaScene} about crowd counting are too rich to elaborate all of them. Next, we briefly review some of them.

\subsection{Counting by detection}  In early researches, many kinds of detection frameworks are proposed in crowd counting. The pivotal features from human body are extracted by the well-trained classifiers such as HOG \cite{dalal2005histograms}, Random Forest \cite{pham2015count}, and Haar wavelets \cite{viola2004robust}, then the classifier outputs the positive samples. The total number of positive samples represents the crowd count. \cite{li2008estimating,zeng2010robust,lin2010shape} complete the crowd counting task by using the aforementioned method. And \cite{Sajid2017Crowd} leverages adaptive thresholds to binarize the image to detect the crowd. They can get a good result in the sparse crowd scene. However, when the crowd density becomes high, some persons are too small to be detected.

\subsection{Counting by regression} In the highly congested crowd scene, the regression-based methods are usually chosen. They have two important stages. The low-level features are first extracted from the input image, then the crowd count is regressed according to the low-level features. To record the space information of the crowd, Lempitsky et al. \cite{lempitsky2010learning} proposed a good method that could learn a linear mapping between local area features and corresponding density maps. Based on Lempitsky's work, Pham et al. \cite{pham2015count} proposed a more applicable method. The linear mapping was replaced with a non-linear mapping based on random forest regression. In \cite{SivaScene}, SVM is proposed to map the features extracted by AdaBoost \cite{Freund1999A} to the crowd counting result.

 
\subsection{Counting by CNN} In recent years, CNN has achieved great success in the computer vision task including crowd counting. Boominathan et al. \cite{boominathan2016crowdnet} combined the deep and shallow CNN to extract large scale features to overcome the scale variation. Zhang et al. \cite{zhang2016single} proposed a three-branch convolutional neural network to encode the density map with different scales. Based on Zhang's work, Sam et al. \cite{sam2017switching} designed a classifier to select the most appropriate branch to generate the density map according to the density variation in each image. Similarly, Hossain et al. \cite{hossain2019crowd} proposed a scale-aware method to make each branch in their model focus on a particular scale. Mahmood et al. \cite{Mahmood2017} designed an Encoder-Decoder CNN to learn the head heatmap where human heads can be segmented. Shen  et al. \cite{shen2018crowd} introduced the scale-consistency model to alleviate the blurry effects in the estimated density map. Sam et al. \cite{sam2018top} designed the TDF-CNN which contains the bottom-up and top-down CNNs to obtain initial accurate predictions. Marsden et al. \cite{marsde2018people} proposed a CNN model that can count the crowd, wildlife, vehicles, and cells. Wang et al. \cite{Wang_2019_CVPR} designed a collector and labeler to generate the crowd data with corresponding labels to reduce the overfitting caused by limited training crowd data. Xu et al. \cite{Xu2019} proposed a segmentation branch based on the probability to guide their framework to encode the final density map. In this paper, we propose a novel segmentation attention mechanism to guide the end-to-end architecture called SegCrowdNet to pay more attention to the human head region. Our segmentation is based on the dice coefficient and we employ the CNN and ground-truth density map to finetune the final density map. Moreover, our proposed SegCrowdNet can adapt the variation of crowd counts by a classification task.


\section{Methodology}


\subsection{SegCrowdNet Architecture}
An overview of the proposed SegCrowdNet can be seen in Fig. 2. The backbone is designed to extract multi-scale features. The proposed SegCrowdNet contains the classification task, the segmentation task, and the density estimation task. 

The configurations of the backbone are shown in Fig. 3. To extract multi-scale features to overcome the scale variations of the crowd, firstly, four different receptive fields are inserted into our system at the beginning of this backbone. Each of them has 16 filters. The input image is mapped at different scales by them synchronously, then the results are fed to the following modules. Secondly, inside this module, we design several 2x2 max-pooling layers to extract multi-scale features. Thirdly, we fuse the feature maps with complementary information from different depths of convolutional layers. Every convolutional layer is followed by ReLU. In order to increase the depth of the network to enhance its learning ability without introducing too many parameters, the parameters of the shared module are shared  to alleviate overfitting caused by excessive parameters. Since the dilated convolution \cite{yu2015multi} can increase the receptive field with fewer parameters, the dilated convolution is widely used in our SegCrowdNet.

In Table 1, it can be observed that the crowd count changes greatly across images in the 'Range' column. For example, the UCF\_CC\_50 dataset only contains 50 images. However, the crowd counts range from 94 to 4,543. So the classification task is employed. The classifier can automatically learn the crowd count distribution to adapt the variation of crowd counts. Inspired by \cite{sindagi2017cnn}, the crowd count of each image in the crowd dataset is quantized into several groups. As shown in Fig 2, the fully connected (FC) layers are connected at the end of the backbone. The two fully connected layers followed by PReLU separately have 64 neurons and 5 neurons. The 5 neurons represent five count groups. To avoid the distortion of images and maintain the original distribution of the crowd, we do not resize the input image. The Spatial Pyramid Pooling (SPP) \cite{he2015spatial} layers are placed between convolutional layers and FC layers. Arbitrarily sized feature map extracted from the input image can be fed to the SPP layer, and the SPP layer produces fixed size outputs to feed the FC layers. In the classification task, the crowd counts of each dataset are classified into five count groups to adapt the variation of crowd counts. We choose to minimize the cross-entropy loss to optimize this process.

Two other tasks are also shown in Fig. 2. In the segmentation task, the segmentation map which adaptively emphasizes the head region and suppresses the non-head region is encoded with the supervision of ground-truth
segmentation. In the density estimation task, the estimated density map (outlined in blue) and the final estimated density map are predicted with the supervision of ground-truth density map. Integrating the final estimated density map can get the crowd count. As shown in Fig 2, we add the segmentation map to the estimated density map (outlined in blue) and the results are fed to the following convolution layers and ReLU layers to automatically encode the final estimated density map. With the guidance of the segmentation map where the human head regions have higher weights, more attention is paid to the human head region in the density estimation task. In the segmentation task, a novel loss based on the dice coefficient is employed. The novel loss and the ground truth of segmentation will be elaborated in Sec 3.2. In the density estimation task, the Euclidean distance loss is employed to optimize the estimated density map and the final estimated density map.



\subsection{Model Learning}\label{sec4}

\noindent{\bfseries Ground truth generation:} The ground-truth density map is extremely important for the supervised methods. For any training image, the 2D point $p$ located at the center of each human head is provided. The ground-truth density map is encoded by employing a normalized Gaussian kernel centered on each point $p$, which is defined as: 
\begin{equation}
 D(c)=\sum _{p\in A _{j}}N(p;\mu,\sigma),
\end{equation} where $c$ represents the pixel location. $A_{j}$ represents a series of 2D points annotated in the image $j$. $N(p;\mu,\sigma) $ represents the normalized Gaussian kernel with mean $\mu=0$ and isotropic variance $\sigma=4$. The window size of the Gaussian kernel is 15$\times$15. We utilize the above simple method to generate the ground-truth density map to ensure that the improvement of the result comes from our novel method instead of the innovation of generating a ground-truth density map.


It is extremely expensive to manually label the ground truth of segmentation, owing to the huge amount of people in datasets. For example, there are 330,165 people in the ShanghaiTech dataset. So a simple but effective method is proposed to encode the ground-truth segmentation map which should have the same foreground and background with the ground-truth density map. First, we construct the ones template that keeps the same size with the Gaussian kernel, which is set to 15$\times$15. Second, the ground truth of segmentation is encoded by pasting the ones template that is centered on each annotated point $p$ on a binary map. They are visualized in Fig. 4. The experiments using different scale templates are conducted on the challenging ShanghaiTech Part\_A. Results are given in Table 2. It can be found that performance with 15$\times$15 is the best. We use the same scale template (15$\times$15) on all four datasets to show the robustness of our proposed method.

\begin{table*}
\caption{Statistics of the training datasets: Num represents the number of images; Range represents the range of crowd count in each image; Average represents the average crowd count; Total represents the total number of labeled people.}
\begin{center}
\label{my-label}
\begin{tabular}{cccccccc|}\hline
Dataset                                & Resolution & Color & Num & \textbf{Range} & Average & Total \\ \hline 

ShanghaiTech Part\_A           & different   & RGB,Grey  & 482  & \textbf{[33, 3139]}  & 501.4   & 241,677 \\ 
ShanghaiTech Part\_B                   & 768 x 1024   & RGB  & 716  & \textbf{[9,  578]}  & 123.6   & 88,488 \\ 
UCF\_CC\_50               & different   & Grey  & 50   & \textbf{[94,  4543]}  & 1279.5   & 63,974 \\ 
WorldExpo’10           & 576 x 720   & RGB  & 3980  & \textbf{[1,  253]}  & 50.2   & 199,923 \\ \hline
\end{tabular}
\end{center}
\end{table*}

\begin{figure}[t]
\begin{center}
   \includegraphics[width=1.0\linewidth]{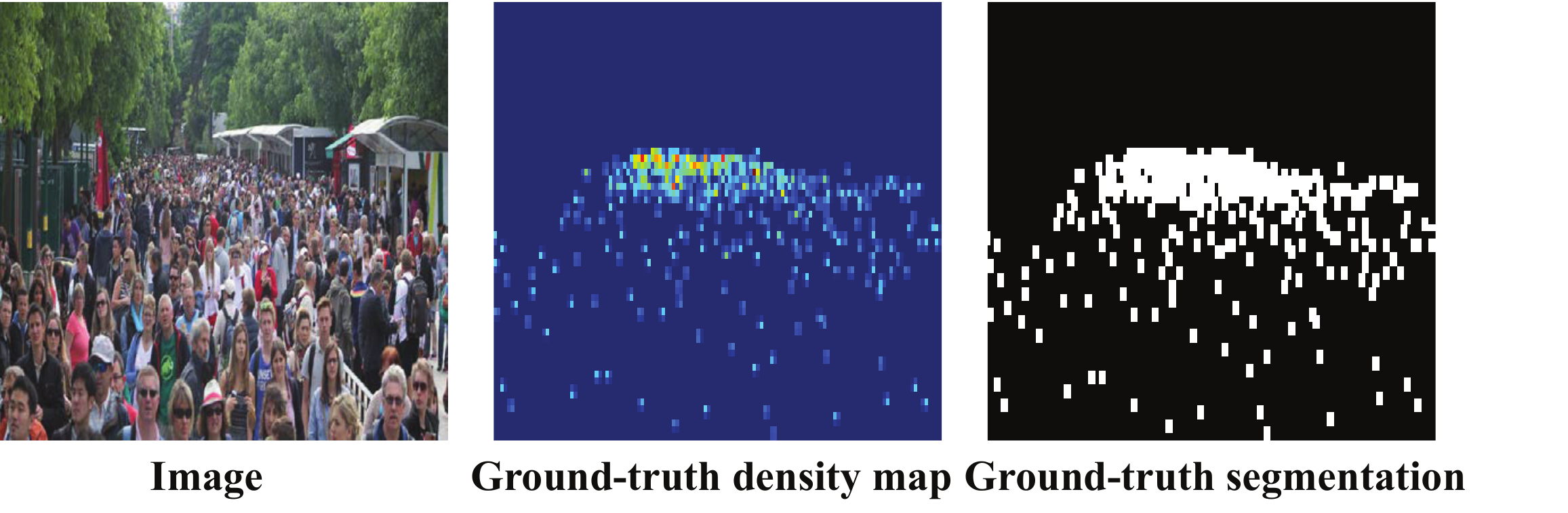}
\end{center}
   \caption{Some visualized results of ground-truth density map and ground-truth segmentation.}
\label{fig:long}
\label{fig:onecol}
\end{figure}


\noindent{\bfseries Model Optimization:} The proposed SegCrowdNet is a multi-task model including classification task, segmentation task, and density estimation task. The multi-task learning \cite{collobert2008unified} can serve as a regularization to alleviate overfitting, which requires our proposed SegCrowdNet to consider every task synchronously, not only one of them. The proposed SegCrowdNet is optimized by minimizing four loss functions in a synergistic manner, including an intermediate supervision loss.
In the density estimation task, the Euclidean loss is utilized to optimize the estimated density map (outlined in blue) and the final estimated density map,

\begin{equation}
L_{int} = \frac{1}{2U}\sum _{i=1}^{U}\left \| \hat{d}_{i} - D_{i} \right \|^{2}_{2},  
\end{equation}

\begin{equation}
L_{den} = \frac{1}{2U}\sum _{i=1}^{U}\left \| \hat{D}_{i} - D_{i} \right \|^{2}_{2},  
\end{equation}
where $\hat{d}_{i} $ represents the estimated density in the intermediate supervision process. $\hat{D}_{i} $ represents the final estimated density. $ D_{i} $ represents the ground-truth density. $U$ represents the number of pixels in the ground-truth density map.

\begin{table}
\begin{center}
\caption {Estimation errors of the template with different sizes on ShanghaiTech \protect\\ Part\_A.}
\label{my-label}
\begin{tabular}{cccccc}
\hline
                            
Template size                       & MAE           & MSE                     \\ \hline

5$\times$5  & 110.1         & 160.7         \\ 
10$\times$10  & 101.3         & 139.0         \\
15$\times$15  & \textbf{68.3}        & \textbf{104.1}         \\
20$\times$20  & 74.8         & 114.5         \\
25$\times$25      &68.5         & 106.7         \\ \hline    
\end{tabular}
\end{center}
\end{table}

In the segmentation task, inspired by \cite{milletari2016v,zettinig2015multimodal}, we introduce a novel loss in crowd counting which is based on the dice coefficient. The loss is optimized to predict the segmentation map for the human head. The dice coefficient $D(\hat{s_{i}},s_{i})$ is between 0 and 1. In the process of optimization, $D(\hat{s_{i}},s_{i})$ is expected to maximize, and the loss $L_{seg}$ is expected to minimize. They are formulated as;
\begin{equation}
D(\hat{s_{i}},s_{i}) = \frac{2\sum_{i}^{U}\hat{s_{i}}s_{i}}{\sum_{i}^{U}\hat{s_{i}}^{2} + \sum_{i}^{U}s_{i}^{2} },
\end{equation}
\begin{equation}
L_{seg} = 1 - D(\hat{s_{i}},s_{i}),
\end{equation}
where U represents the total number of pixels in the ground truth segmentation map. $s_{i}$ represents the $i^{th}$ value in the ground truth segmentation map. $\hat{s_{i}}$ represents the $i^{th}$ value in the predicted segmentation map.

In the classification task, the crowd counts are quantized into five groups. For example, if the crowd counts of a dataset range from 1 to 500 and they are quantized into five classes, the images with populations between 1 and 100 belong to the first class, and the images with populations between 401 and 500 belong to the fifth class. The cross-entropy loss function is utilized, 
\begin{equation}
L_{cla} = -\frac{1}{M}\sum ^{M}_{a=1}\sum ^{K}_{b=1}\left [ \left ( y^{a}_{b} \right )log (\hat{y}^{a}) \right ],
\end{equation}
where $M$ represents the total number of training samples. $K$ represents the total number of classes. $y^{a}_{b}$ represents the ground-truth class. $\hat{y}^{a}$ represents the output of classification.
The final weighted loss is given by equation (7). Empirically, we set $ \lambda_{1}$ as 0.01. 
\begin{equation}
L_{fin} =  L_{den} + L_{int}+ L_{seg} + \lambda_{1} L_{cla},
\end{equation}


\begin{figure*}[t]
\begin{center}
    \includegraphics[width=1.0\linewidth]{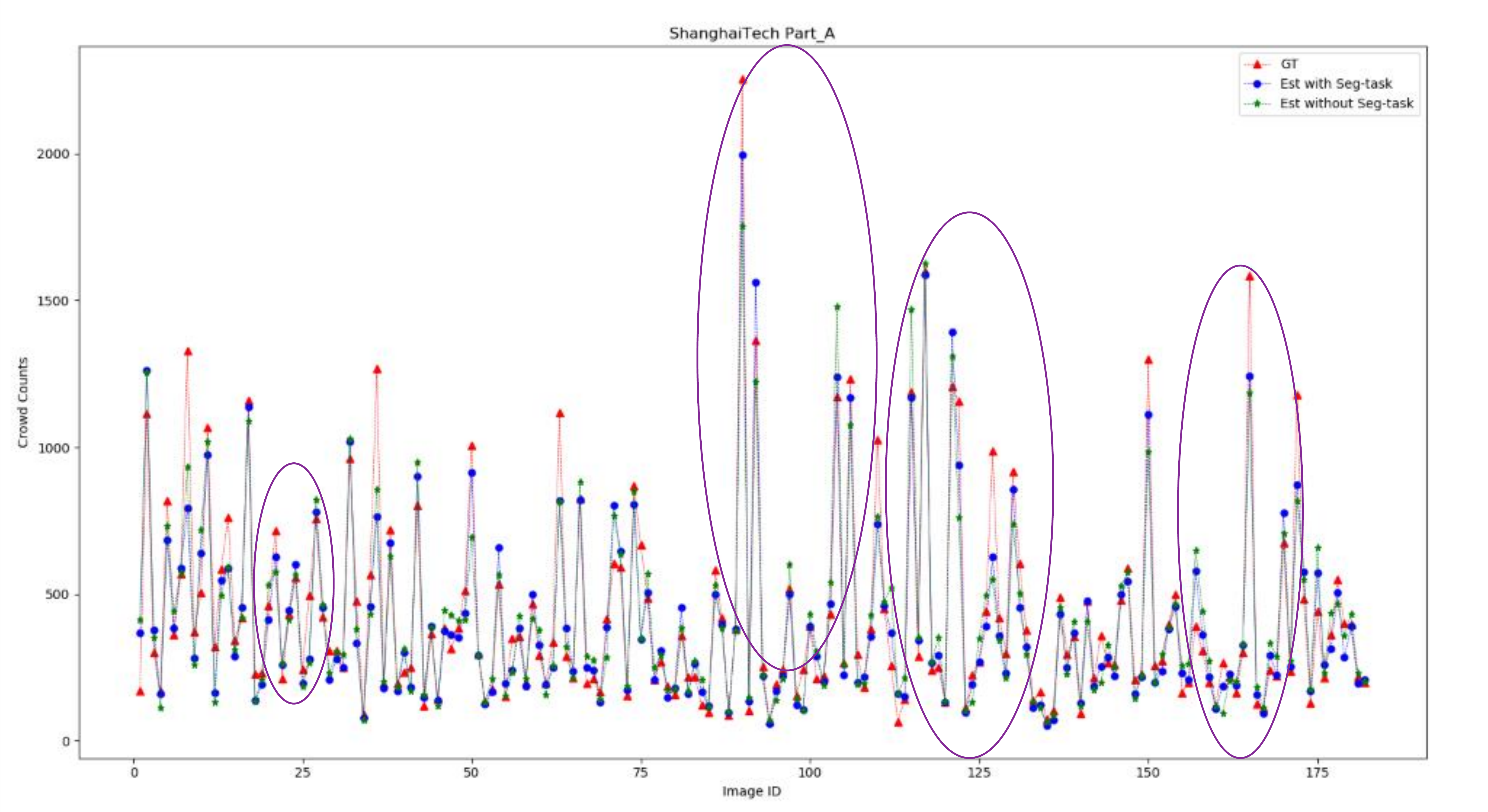}
\end{center}
   \caption{The crowd count comparisons of different configurations about the Seg-task on ShanghaiTech Part\_A. X-axis: the ID of images, Y-axis: the crowd counts of images.}
\label{fig:long}
\label{fig:onecol}
\end{figure*}

\begin{figure}[H]
\begin{center}
   \includegraphics[width=1.0\linewidth]{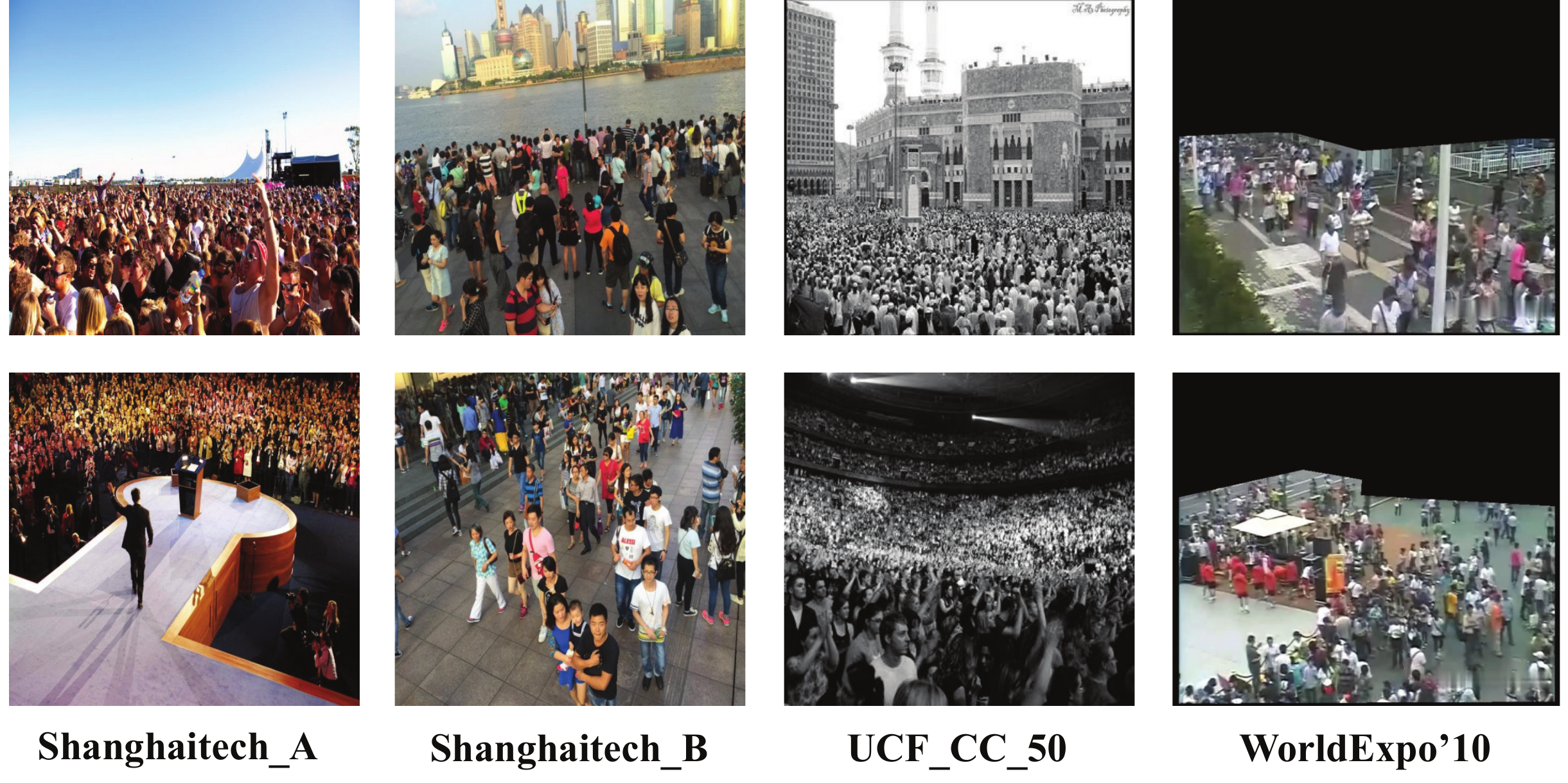}
\end{center}
   \caption{Some example frames from training datasets.}
\label{fig:long}
\label{fig:onecol}
\end{figure}

\section{Experiments}
\subsection{Experiment setting}
 We conduct extensive experiments on four challenging datasets. The statistics of the four challenging datasets are summarized in Table 1. Some example frames from these four datasets are shown in Fig. 6. In the process of creating training data, 9 patches are cropped from each original image randomly, and the size of each patch is 1/4 of the original image. Horizontal flipping and noise addition are also used in the training data. We utilize Adam optimization with a fixed constant momentum of 0.9 to train our model. And we set the learning rate as 1e-6.

\begin{table}
\begin{center}
\caption {Estimation errors of different configurations about the Cla-task\protect\\ on ShanghaiTech Part\_A.} 

\label{my-label}
\begin{tabular}{cccccc}
\hline
                            
Method                       & MAE           & MSE                     \\ \hline

Without Cla-task  & 80.7         & 124.0         \\ 
With Cla-task      &\textbf{68.3}         & \textbf{104.1}         \\ \hline    
\end{tabular}
\end{center}

\end{table}

\begin{table}
\begin{center}
\caption {Estimation errors of different categories about the crowd counts\protect\\ on ShanghaiTech Part\_A.} 

\label{my-label}
\begin{tabular}{cccccc}
\hline
                            
categories                      &3           & 5    & 7   & 10   & 15              \\ \hline

MAE           &71.3 & \textbf{68.3}   &77.8   &78.4   &70.4        \\ \hline    
\end{tabular}
\end{center}

\end{table}

\begin{table}
\centering
\caption {Estimation errors of different configurations about the Seg-task\protect\\ on ShanghaiTech Part\_A.} 

\label{my-label}
\begin{tabular}{cccccc}
\hline
                            
Method                       & MAE           & MSE                     \\ \hline

Without Seg-task  & 85.8         & 128.6         \\ 
With Seg-task      &\textbf{68.3}         & \textbf{104.1}         \\ \hline    
\end{tabular}

\end{table}

 \subsection{Evaluation Metric} 
The mean absolute error (MAE) and the mean squared error (MSE) are widely used to measure the crowd counting error on different methods \cite{zhang2016single,sam2017switching,sam2018top,Wang_2019_CVPR}. They are defined as follows:

\begin{equation}
MAE = \frac{1}{V}\sum _{i=1}^{V}\left | z_{i}- \hat{z_{i}} \right |,                 
\end{equation}

\begin{equation}
MSE = \sqrt{\frac{1}{V}\sum ^{V}_{i=1}\left ( z_{i} - \hat{z_{i}} \right )^{2}}.  
\end{equation}
Where $z_{i}$ represents the ground-truth crowd count. $ \hat{z_{i}}$ represents the estimated crowd count. $V$ represents the number of test images,

\subsection{Ablation study using ShanghaiTech Part\_A}
To verify the effectiveness of each task in the proposed SegCrowdNet, we perform ablation studies on ShanghaiTech Part\_A \cite{zhang2016single} which is a large scale and high-density dataset, the dataset contains 482 images with 241,667 annotated persons. In other datasets, similar results can be observed.

To investigate the ability of this classification task where the crowd count of each image is classified, we conduct a set of comparative experiments. In one experiment, the classification task can function normally. In the second experiment, the classification task is removed. The estimation errors are given in Table 3. We can observe that the performance with Cla-task is better, with MAE/MSE 12.4/19.9 lower than that without Cla-task. We think that the proposed SegCrowdNet can learn the crowd count distribution to adapt the variation of crowd counts, which contributes to reducing the crowd counting errors. In the proposed SegCrowdNet, we classify the crowd counts of images in each dataset into five categories according to extensive experiments. From Table 4, we find that our proposed SegCrowdNet with five categories perform best. We think that the crowd counting data are limited and five categories are most suitable. Last but not least, we can further observe that adopting the classification task with different categories in Table 4, the MAE is further alleviated compared with the performance without Cla-task in Table 3, which further indicates that the Cla-task can assist in reducing the estimation error of crowd counting.

\begin{figure*}[t]
\setlength{\abovecaptionskip}{0.05cm}
\setlength{\belowcaptionskip}{-0.3cm}

\begin{center}
   \includegraphics[width=1.0\linewidth]{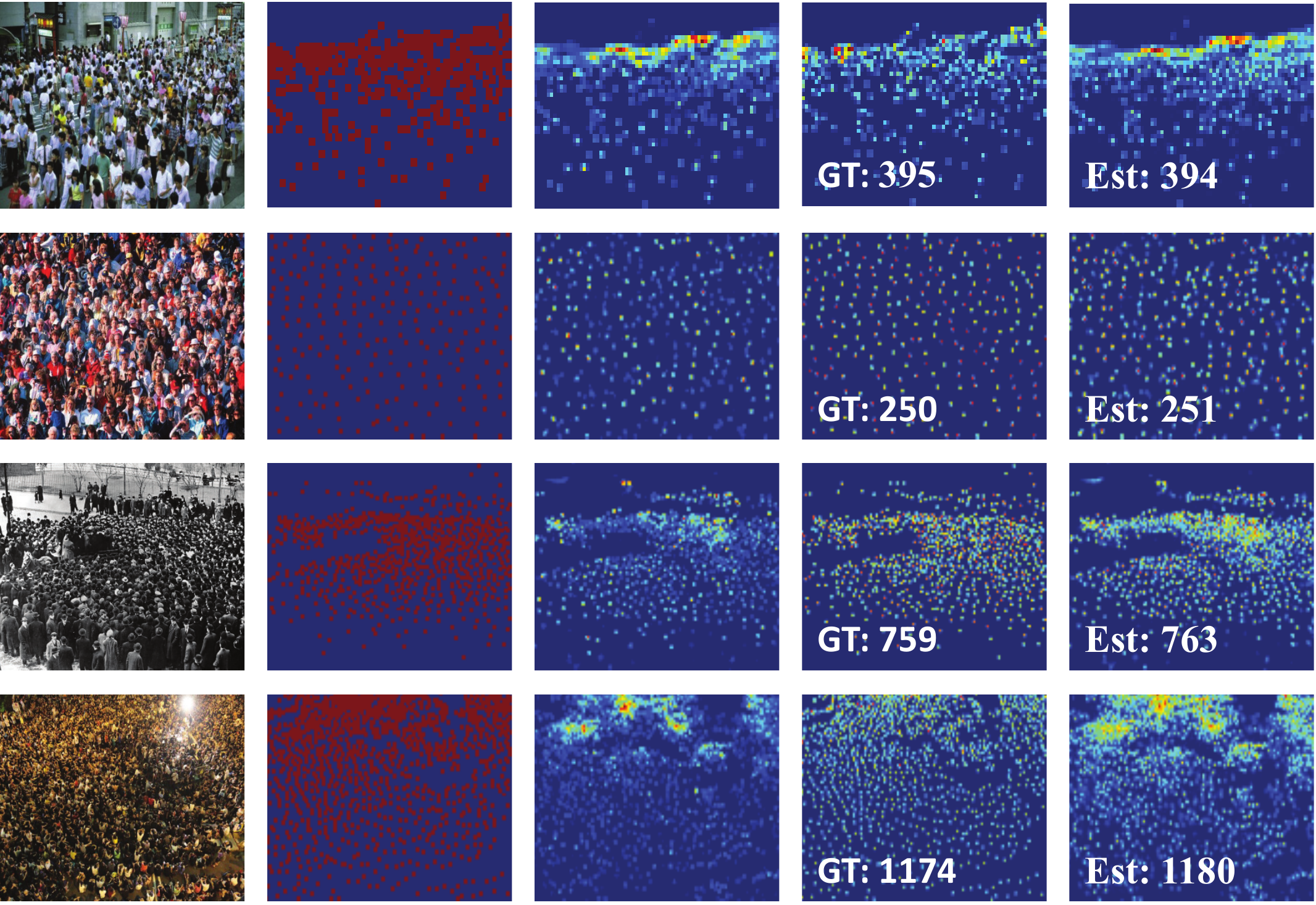}
\end{center}
   \caption{Several visualization results of the ShanghaiTech Part\_A. Top Column: Some input images from the ShanghaiTech Part\_A dataset. Second Column: The ground-truth segmentation. Third Column: The predicted segmentation map. Fourth Column: The ground-truth density map. Fifth Column: The final estimated density map. GT: The crowd count of the ground-truth density map. Est: The crowd count of the estimated density map.  }
\label{fig:long}
\label{fig:onecol}
\end{figure*}

The proposed segmentation attention mechanism is the most important idea in this paper. It is included in the segmentation task. To demonstrate the effectiveness of the segmentation task, we also conduct a set of comparative experiments. In one experiment, the segmentation task can function normally. In the second experiment, the segmentation task is removed from our proposed SegCrowdNet and the segmentation attention mechanism is subsequently removed. The qualitative results are illustrated in Table 5. We can observe that the SegCrowdNet greatly reduces the error of crowd counting with the guidance of segmentation results, with MAE/MSE 17.5/24.5 lower than that without Seg-task, which reveals the power of the segmentation attention mechanism. In Fig. 5, we demonstrate the estimated crowd count of each image in the comparative experiment. It can be observed that the blue circle (predicted crowd counts with the guidance of the Seg-task) is closer to the red triangle (actual crowd counts) than the green pentagram (predicted crowd counts without Seg-task) in most images, which indicates that the segmentation attention mechanism plays a vital role in alleviating the errors of crowd counting.

 We visualize the effectiveness of the Seg-task in Fig. 7. In the third column, it can be observed that the human head regions are highlighted and the non-head regions are suppressed excellently in the estimated segmentation map, which is important to guide our proposed SegCrowdNet to focus on the human head region. In the fourth and fifth columns, we can observe that with the guidance of the segmentation results, the proposed SegCrowdNet pays more attention to the human head region and the crowd counts are predicted beautifully. Similar results can be found in Fig. 8. From these results, we can see that every task in the proposed SegCrowdNet is necessary. They work collaboratively to better accomplish the task of crowd counting.


\begin{figure*}[t]

\begin{center}
   \includegraphics[width=1.0\linewidth]{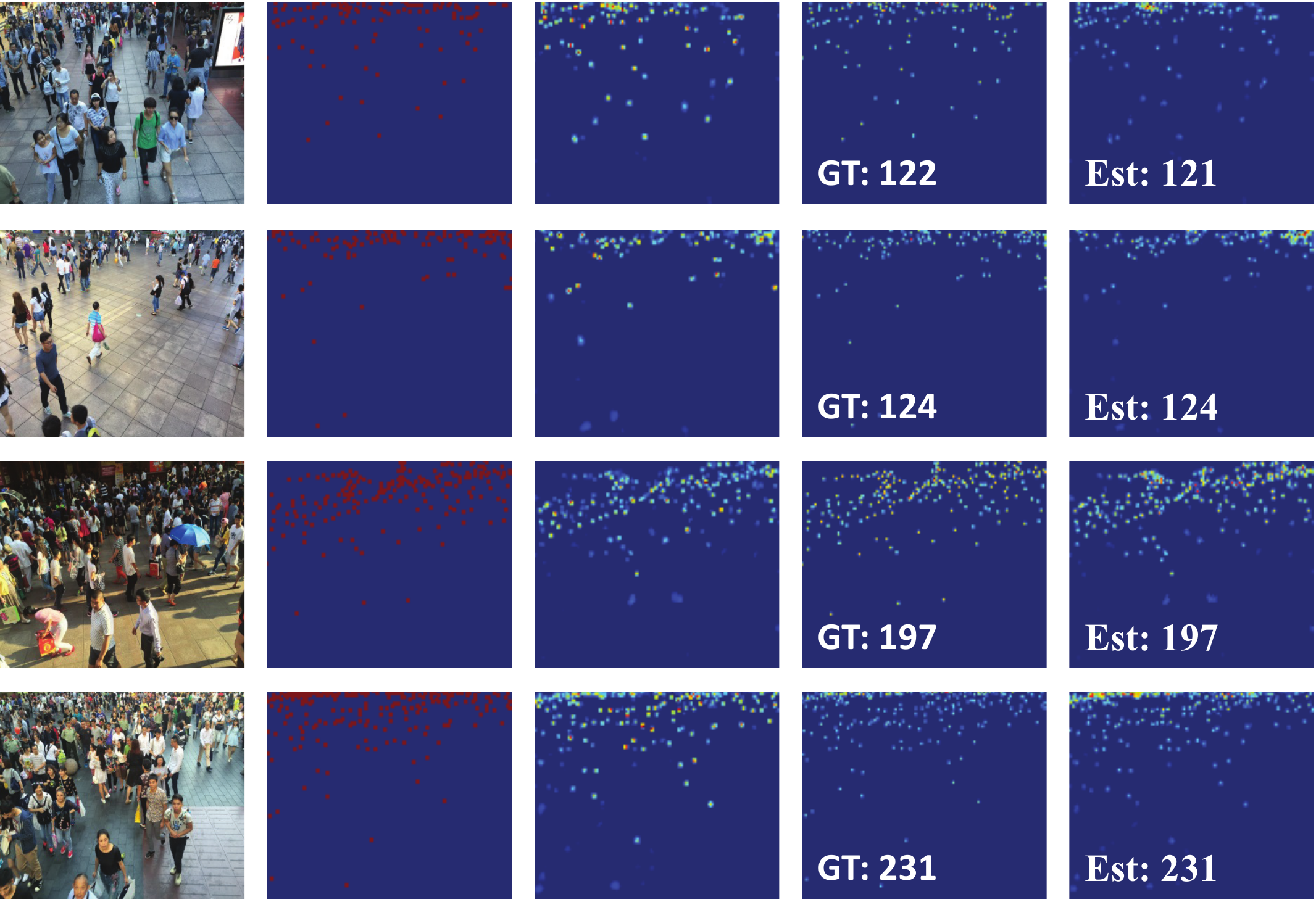}
\end{center}
   \caption{Several visualization results of the ShanghaiTech Part\_B. Top Column: Some input images from the ShanghaiTech Part\_B dataset. Second Column: The ground-truth segmentation. Third Column: The predicted segmentation map. Fourth Column: The ground-truth density map. Fifth Column: The final estimated density map. GT: The crowd count of the ground-truth density map. Est: The crowd count of the estimated density map.  }
\label{fig:long}
\label{fig:onecol}
\end{figure*}

\subsection{ShanghaiTech dataset}   
The ShanghaiTech dataset \cite{zhang2016single} is highly challenging because of its large-scale crowds. It contains 1,198 annotated images with a total number of 330,165 persons. So far the dataset contains the richest annotated crowd information and it is composed of two parts: Part\_A with 300 training images and 182 test images, Part\_B with 400 training images and 316 test images. Part\_A consists of the high-density crowd images downloaded from the internet. Part\_B consists of the relatively low-density crowd images collected from Shanghai streets.

 As shown in Table 6, we compare our proposed method with the other fourteen state-of-the-art methods. In the LBP+RR method, the LBP features were extracted by hand to regress the crowd count. Owing to the scale variation in the crowd, the work of MCNN \cite{zhang2016single}, CAFN \cite{fan2018cafn}, TDF-CNN \cite{sam2018top}, Switch-CNN \cite{sam2017switching}, and CP-CNN  \cite{sindagi2017generating} employed multi-column CNNs to extract the multi-scale features to cope with the scale variation in the crowd. Instead of using the multi-column CNN, in \cite{hossain2019crowd}, Hossain et al. proposed the scale-aware attention network to encode the multi-scale density maps. In \cite{babu2018divide}, IG-CNN adapted the multi-scale crowd scenes by increasing its capacity. The work of Cascaded-MTL \cite{sindagi2017cnn}, Marsden et al.\cite{marsde2018people}, DecideNet+R3 \cite{liu2018decidenet}, and DDCN \cite{wang2019removing} introduced the multi-task learning to assist in reducing the count estimation errors. The work of SE Cycle GAN \cite{Wang_2019_CVPR} designed the data collector and labeler to generate the crowd data with the corresponding labels to alleviate the overfitting caused by limited data for training. In \cite{shi2018crowd}, the deep negative correlation learning was utilized to extract the general feature representation. The estimation errors are illustrated in Table 6. It can be observed that, firstly, the performance of deep learning is far better than the hand-crafted features. Secondly, lots of efforts are done to extract the multi-scale features of the crowd, which is very important to reduce the estimation errors to some extent. Finally, our proposed SegCrowdNet achieves the best performance in MAE and MSE on both datasets.

\begin{table}[]
\begin{center}
\centering
\caption{Estimation errors on the ShanghaiTeach dataset.}
\label{my-label}
\begin{tabular}{cccccc}
\hline
Method                & \multicolumn{2}{c}{Part\_A}   & \multicolumn{2}{c}{Part\_B}  \\ 
                    & MAE           & MSE            & MAE           & MSE           \\ \hline

LBR + RR  & 303.2         & 371.0          & 59.1          & 81.7          \\ 
SE Cycle GAN \cite{Wang_2019_CVPR}        & 123.4         & 193.4         & 19.9         & 28.3          \\ 
MCNN \cite{zhang2016single}       & 110.2         & 173.5          & 26.4          & 41.3          \\ 
Cascaded-MTL \cite{sindagi2017cnn}   & 101.3         & 152.4          & 20.0          & 31.1          \\ 
CAFN \cite{fan2018cafn}   &100.8        & 152.3          & 21.5 & 33.4            \\ 
TDF-CNN \cite{sam2018top}        & 97.5          & 145.1          & 20.7          & 32.8          \\ 
Switch-CNN \cite{sam2017switching} & 90.4          & 135.0          & 21.6          & 33.4          \\ 

Marsden et al.\cite{marsde2018people}                 & 85.7          & 131.1          & 17.7          & 28.6          \\ 

DecideNet+R3 \cite{liu2018decidenet}       &- \footnotemark[1]     & -         & 20.8          & 29.4          \\ 

CP-CNN  \cite{sindagi2017generating}       &   73.6          & 106.4 & 20.1         & 30.1          \\  
D-ConvNet-V1 \cite{shi2018crowd}   &73.5        &112.3        & 18.7 & 26.0            \\ 
Hossain et al. \cite{hossain2019crowd}       &-         & -         & 16.9          & 28.4          \\ 
IG-CNN \cite{babu2018divide}  &72.5        &118.2        & 13.6 & 21.2            \\ 
DDCN \cite{wang2019removing}  &71.5        &110.4        & 13.8 & 20.1            \\


OURS                         &\textbf{68.3}        & \textbf{104.1}      &\textbf{ 12.1}          & \textbf{19.3 }       \\ \hline
\end{tabular}
\end{center}
\end{table}
\footnotetext[1]{'-' represents that the result of this dataset is not reported in that paper.}

\subsection{UCF\_CC\_50 dataset} 
The UCF\_CC\_50 dataset \cite{idrees2013multi} contains 50 crowd images collected from the internet. The number of people in each image ranges from 94 to 4,543. There are only 50 images in this dataset. We follow the standard protocol discussed in \cite{idrees2013multi}. The 5-fold cross-validation is utilized to evaluate the performance on this dataset.


\begin{table*}[t]
\begin{center}
\caption{Estimation errors on the WorldExpo'10 dataset.}

\scalebox{1.0}{
\label{my-label}
\begin{tabular}{ccccccc}
\hline
Method                                & S1 & S2 & S3 & S4 & S5 & Average \\ \hline
LBP+RR                                & 13.6   & 58.9  & 37.1  & 21.8  & 23.4   & 31.9 \\ 
SE Cycle GAN \cite{Wang_2019_CVPR}                   & 4.3   & 59.1  & 43.7  & 17.0  & 7.6   & 26.3 \\
MCNN \cite{zhang2016single}                   & 3.4   & 20.6  & 12.9  & 13.0  & 8.1   & 11.6 \\
TDF-CNN \cite{sam2018top}                & 2.7   & 23.4  & 10.7  & 17.6  & 3.3   & 11.5  \\
IG-CNN \cite{babu2018divide}                & 2.6   & 16.1  & 10.2   & 20.2  & 7.6   & 11.3  \\
DDCN \cite{wang2019removing}       & 4.8   & 16.2  &12.4   & 10.9  & 4.9   & 9.8  \\   
Switch-CNN \cite{sam2017switching}        & 4.4   & 15.7  & 10.0  & 11.0 & 5.9   & 9.4  \\
DecideNet \cite{liu2018decidenet}       & 2.0   & 13.1  &\textbf{ 8.9}   & 17.4  & 4.8   & 9.2  \\
D-ConvNet-V1 \cite{shi2018crowd}       & \textbf{1.9}   & 12.1  & 20.7   & \textbf{8.3}  & 2.6   & 9.1  \\ 
CP-CNN \cite{ sindagi2017generating}      & 2.9   & 14.7  & 10.5  &  10.4 & 5.8   & 8.9 \\ 
OURS                                  & 2.3   & \textbf{11.9}  & 11.8  & 11.2  &\textbf{ 2.4}   &\textbf{7.9}  \\ \hline
\end{tabular}}
\end{center}
\end{table*}

\begin{table}[]
\begin{center}
\centering
\caption{Estimation errors on the UCF\_CC\_50 dataset.}
\label{my-label}
\begin{tabular}{ccccc}
\hline
\multicolumn{1}{c}{Method}                             & \multicolumn{1}{c}{MAE}            & \multicolumn{1}{c}{MSE}            \\ \hline
\multicolumn{1}{c}{Rodriguez et al.  \cite{rodriguez2011density} }         & \multicolumn{1}{c}{655.7}          & \multicolumn{1}{c}{697.8}          \\ 
\multicolumn{1}{c}{Lempitsky et al.     \cite{idrees2013multi} }        & \multicolumn{1}{c}{419.5}          & \multicolumn{1}{c}{541.6}          \\ 

\multicolumn{1}{c}{MCNN \cite{zhang2016single}}                & \multicolumn{1}{c}{377.6}          & \multicolumn{1}{c}{509.1}          \\ 
\multicolumn{1}{c}{SE Cycle GAN \cite{Wang_2019_CVPR}}              & \multicolumn{1}{c}{373.4}          & \multicolumn{1}{c}{528.8} \\ 
\multicolumn{1}{c}{TDF-CNN \cite{sam2018top}}              & \multicolumn{1}{c}{354.7}          & \multicolumn{1}{c}{491.4} \\ 
\multicolumn{1}{c}{Cascaded-MTL \cite{sindagi2017cnn}}              & \multicolumn{1}{c}{322.8}          & \multicolumn{1}{c}{397.9} \\ 
\multicolumn{1}{c}{Switch-CNN \cite{sam2017switching}}              & \multicolumn{1}{c}{318.1}          & \multicolumn{1}{c}{439.2} \\ 
\multicolumn{1}{c}{CAFN \cite{fan2018cafn}}              & \multicolumn{1}{c}{305.3}          & \multicolumn{1}{c}{429.4} \\ 

\multicolumn{1}{c}{CP-CNN \cite{sindagi2017generating}}              & \multicolumn{1}{c}{295.8}          & \multicolumn{1}{c}{\textbf{320.9}} \\ 
\multicolumn{1}{c}{IG-CNN \cite{babu2018divide}}              & \multicolumn{1}{c}{291.4}          & \multicolumn{1}{c}{349.4}          \\ 
\multicolumn{1}{c}{D-ConvNet-V1 \cite{shi2018crowd}}        & \multicolumn{1}{c}{288.4}          & \multicolumn{1}{c}{404.7}          \\ 
\multicolumn{1}{c}{DDCN \cite{wang2019removing}}        & \multicolumn{1}{c}{286.2}          & \multicolumn{1}{c}{479.6}          \\ 
\multicolumn{1}{c}{Hossain et al. \cite{hossain2019crowd}} & \multicolumn{1}{c}{271.6}          & \multicolumn{1}{c}{391.0}          \\ 

\multicolumn{1}{c}{OURS}                            & \multicolumn{1}{c}{\textbf{233.6}} & \multicolumn{1}{c}{352.6}          \\ \hline
\end{tabular}
\end{center}

\end{table}

 In Table 8, we compare our method with other recent state-of-the-art methods. The hand-crafted features were extracted in \cite{rodriguez2011density,idrees2013multi}. The methods which have been compared on the ShanghaiTech are still compared in this dataset. The detailed comparison results are shown in Table 8. In the same way, it can be observed that all of the CNN-based methods outperform the methods of traditional feature extraction \cite{rodriguez2011density,idrees2013multi} significantly. The proposed SegCrowdNet leads the best performance in MAE. More specifically, its MAE is 38.0 lower than the second best method. However, the MSE which indicates the robustness of a method is not the best. We think that the root of causing this result lies in the limited data, as there are only 50 images in this dataset.

\subsection{WorldExpo’10 dataset}

The WorldExpo’10 dataset \cite{zhang2015cross} contains 1,132 video sequences that are captured from 108 scenarios and it consists of 3,982 annotated frames with a size of 576$\times$720. The frames with the region of interest (ROI) are divided into a training set with 3380 frames and a test set with 600 frames. The test set consists of 5 five different scenes (S1-S5) and each scene contains 120 frames. For fair comparisons, we followed the methods in \cite{zhang2015cross} to generate the density map.
 

The results of recent state-of-the-art methods are summarized in Table 7. The MAE is utilized to evaluate these methods. We can observe that the CNN-based methods are still superior to traditional methods. Our proposed SegCrowdNet obtains the best result on the S2, S5, and Average. While it achieves comparable performance on the other three scenes. By reviewing the test images of these three scenes, we find that some people are majorly gathered in these three scenes, which is challenging for our proposed SegCrowdNet to further improve their accuracy.

\section{Conclusions}

In this paper, an end-to-end architecture named SegCrowdNet is designed in crowd counting. We propose a novel segmentation attention mechanism to guide our SegCrowdNet to pay more attention to the human head regions. The proposed SegCrowdNet can also automatically adapt the variation of crowd counts by optimizing a classification task. Moreover, the proposed novel four-loss optimization improves the generalization ability of the SegCrowdNet. We verify our method on four popular crowd counting datasets (ShanghaiTech Part\_A dataset, ShanghaiTech Part\_B dataset, UCF\_CC\_50 dataset, and WorldExpo’10 dataset). Extensive experimental results demonstrate that our proposed method outperforms many state-of-the-art methods.


\section{Acknowledgments}\label{sec7}

This work was supported by the National Natural Science Foundation of China (No.61472393).

\noindent{\bfseries \textcolor{red}{Please cite
@article\{chen2021crowd,
  title=\{Crowd counting with segmentation attention convolutional neural network\},
  author=\{Chen, Jiwei and Wang, Zengfu\},
  journal=\{IET Image Processing\},
  volume=\{15\},
  number=\{6\},
  pages=\{1221--1231\},
  year=\{2021\},
  publisher=\{Wiley Online Library\}
\}
}
}


\bibliographystyle{iet}
\bibliography{mulu_13_1}

\end{document}